\documentclass[10pt,twocolumn,letterpaper]{article}

\usepackage{cvpr}
\usepackage{times}
\usepackage{epsfig}
\usepackage{graphicx}
\usepackage{amsmath}
\usepackage{amssymb}
\usepackage{makecell}
\usepackage{caption}

\usepackage[pagebackref=true,breaklinks=true,letterpaper=true,colorlinks,bookmarks=false]{hyperref}

\cvprfinalcopy 
\def\cvprPaperID{****} 

\ifcvprfinal\fi
\begin{document}

\title{
Optimizing Inference in Transformer-Based Models
}

\author{Daniel Schlabig\\
{\tt\small dschlabig@gatech.edu}
\and
Nguyen Duong\\
{\tt\small nguyen.duong@gatech.edu}
\and
Prasad Ganesan\\
{\tt\small pganesan8@gatech.edu}
\and
Siu Hang Ho\\
{\tt\small williamsiuhang@gatech.edu}
}

\maketitle

\begin{abstract}
   Efficient inference is a critical challenge in deep generative modeling, particularly as diffusion models grow in capacity and complexity. While increased complexity often improves accuracy, it raises compute costs, latency, and memory requirements. This work investigates techniques such as pruning, quantization, knowledge distillation, and simplified attention to reduce computational overhead without impacting performance. The study also explores the Mixture of Experts (MoE) approach to further enhance efficiency. These experiments provide insights into optimizing inference for state-of-the-art Fast Diffusion Transformer (fast-DiT) model.
\end{abstract}

\section{Introduction / Background / Motivation}

Although diffusion transformers~\cite{peebles2023scalablediffusionmodelstransformers} deliver high accuracy and image quality, their heavy computational demands impede real-time and low-resource deployment. This paper studied methods to reduce the computational cost of deep generative models during inference by measuring changes in compute cost and quality compared to a selected base model. 

Faster inference is important for diffusion models to make high-quality image generation faster and less expensive, lowering the barriers to deploying advanced image generation in a wide range of applications and industries, from mobile devices to real-time systems. Applications include real-time, interactive media and personalized services, scalable, automated workflows in industry, and more accessible state-of-the-art models for hobbyists and researchers, democratizing access across professional and personal environments.

Our research explores various techniques for model and inference optimization, examining their impact on compute cost and image quality. Prior work for each approach is briefly discussed below.

\textbf{Pruning.} Pruning reduces memory requirements and computational complexity of deep learning networks by removing less critical layers or increasing parameter sparsity. Unstructured L2 norm pruning~\cite{molchanov2017pruning} removes weights with the smallest L2 norms, adding sparsity without altering model architecture, making it ideal for fine-grained compression of large models.

\textbf{Quantization.} Quantization reduces the precision of model weights and activations (typically from 32-bit floating-point to 8-bit integers), significantly reducing model size and memory usage while aiming to preserve performance. Recent work shows that quantization techniques can be successfully extended to large-scale transformer and generative models \cite{dettmers2022llmint88bitmatrixmultiplication}. Common approaches include post-training quantization (PTQ), which requires no retraining, and quantization-aware training (QAT), which incorporates quantization during model training \cite{jacob2017quantizationtrainingneuralnetworks,han2016deepcompressioncompressingdeep}. Our work focuses on post-training quantization as a zero-training-cost approach to model compression, making it particularly attractive for large diffusion models where retraining is expensive.

\textbf{Attention.} In transformer models, scaled dot-product attention complexity scales quadratically with token sequence length $N$ and linearly in hidden layer size $C$ ($\mathcal{O} (N^2C)$) \cite{vaswani2023attentionneed}. Reducing or eliminating the quadratic dependence is the focus of much prior work. Methods include neighborhood or window-based approaches to 'localize' most attention \cite{hassani2023neighborhoodattentiontransformer}{\cite{liu2021swintransformerhierarchicalvision}}, and "linearizing" methods \cite{han2023flattentransformervisiontransformer} \cite{pu2024efficientdiffusiontransformerstepwise} that seek to avoid multiplying $QK^T$, eliminating the $N^2$ dependency by calculating $K^TV$ first  at $\mathcal{O} (N^2C)$. 

Others have applied multiple queries to the same keys and values \cite{ainslie2023gqatraininggeneralizedmultiquery} described as grouped-query attention or multi-query attention (when a single K-V head is used). These methods aim to reduce memory bandwidth in cases where memory is the inference bottleneck. 

These methods may be less relevant to smaller models applied to low resolution images, leading us to explore a range of attention simplification approaches.

\textbf{Knowledge Distillation.} Knowledge distillation transfers knowledge from a larger "teacher" model to a smaller "student" model. The student learns to mimic the teacher's behavior while using fewer parameters and computational resources. This technique has been effective in various domains, including image classification \cite{hinton2015distillingknowledgeneuralnetwork}, natural language processing \cite{sanh2020distilbertdistilledversionbert}, and more recently in diffusion models \cite{salimans2022progressivedistillationfastsampling}.

By learning from soft predictions instead of only hard labels, the student gains richer supervision during training. This can improve generalization and enable the student to reach accuracy close to that of the larger teacher model.

\textbf{Mixture of Experts.} Mixture of Experts (MoE) is an architectural optimization technique where for each input, only a subset of specialized expert subnetworks is activated based on token routing. Instead of evaluating the entire model for every input, only a few experts process each token, significantly reducing computational cost while preserving or even increasing model capacity and expressiveness.

When applied to Diffusion Transformers, the MoE mechanism typically replaces the standard MLP layers inside the Transformer blocks with sparse MoE layers. Each token is routed by a lightweight gating network to a small number of experts, enabling the model to scale up in total parameters without proportionally increasing FLOPs or memory usage. This allows larger, more powerful diffusion models. For example, DiT-MoE \cite{fei2024scalingdiffusiontransformers16} scales to 16.5 billion parameters while only activating 3.1 billion per forward pass.

\textbf{Dataset}. ImageNet \cite{ImageNet256} is a large-scale dataset of images organized in a hierarchical ontology. We used a pre-processed subset of 540,000 256x256 images grouped into 1000 classes from Kaggle \cite{kaggle}. These images had been cropped to square and organized into a flat folder structure by class. For our experiments, we randomly sampled 200 classes (350-650 images per class) for approximately 110,000 total images, each with color or grayscale content and a class label. The data source we used was not split into train-validation-test groups. However, since our experiments were generative rather than discriminative (i.e. not classification), we did not split the dataset when training our images.

\section{Approach}

\textbf{Base model selection}. We used the fast-DiT \cite{jin2024fast} re-implementation of DiT as our base diffusion transformer. Its improvements over the original DiT include saving the latent features of the variational autoencoder (VAE) rather than re-computing at each iteration, and quantizing the model parameters as fp16 without changing the original architecture. We compared qualitative performance and training speed of published model parameters B/2, B/4, S/2, and S/4 (each with depth=12, hidden\_size=768, num\_heads=12 for B, hidden\_size=384, num\_heads=6 for S, patch\_size=2 or 4 based on name). We also trialed a smaller model size we named XS with depth=6, hidden\_size=256, num\_heads=4. We selected S/2 as based model based on it producing reasonable quality images with tractable training times for this project. Also, S/2 has a more balanced token sequence length ($N$=256) and hidden dimension size ($C$=384), which is closer to the ratio seen with high resolution image generation using the DiT XL/2 model.

\textbf{Pruning.} We applied L2 norm pruning to reduce the number of attention heads in the S/2 model. For each attention layer, we computed the L2 norm of each head's Q, K, and V weight slices, then summed them to calculate a score for each attention head. We then kept the top-$k$ scoring heads and masked the rest by setting their weights and biases to zero. $k$ was varied between $[2, 5]$.

\textbf{Quantization.} PyTorch's native quantization tools only support CPU inference~\cite{pytorch_quantization_doc}, so we needed an alternative approach for fair GPU-based comparisons with our baseline fast-DiT models. To address this, we implemented TorchAO's \texttt{Int8WeightOnlyConfig()}~\cite{torchao}, which provides experimental support for static int8 weight-only quantization on CUDA through quantized tensor subclasses. Only the model weights were quantized; activations remained in float precision during inference. The quantization function applies an affine transformation:
\begin{equation}
q = \text{round}(x/S) + Z
\end{equation}
where $S$ is a scaling factor and $Z$ is a zero-point, to map real-valued weights into 8-bit integers.

While TorchAO enabled successful quantization, its CUDA backend remains experimental and unoptimized, which likely explains the observed decrease in throughput despite theoretical efficiency gains. Nonetheless, the quantization process integrated cleanly into our training pipeline and enabled efficient model compression across four DiT variants (S/2, S/4, XS/2, XS/4).

\begin{figure}[t]
\begin{center}
\includegraphics[width=1.0\linewidth]{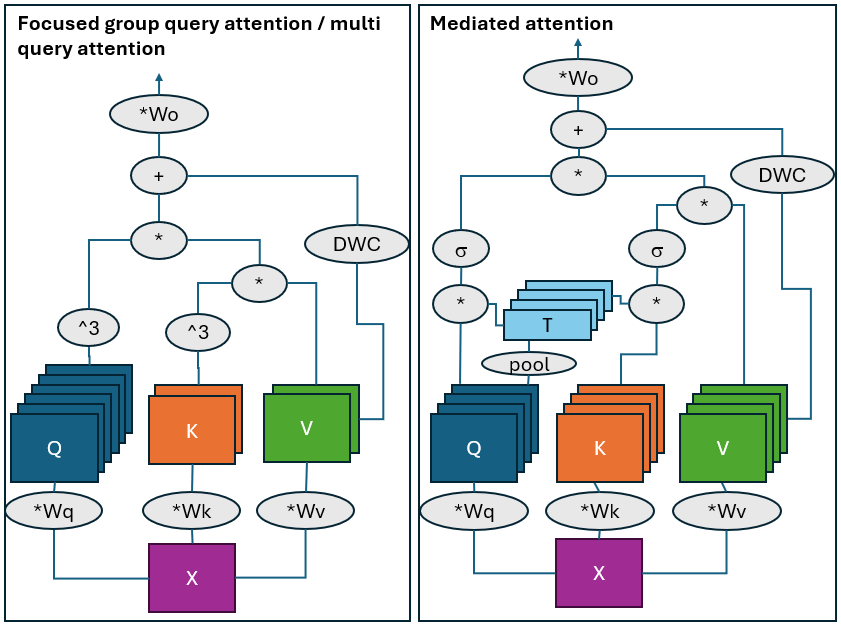}
\end{center}
   \caption{Computation graphs for (left) focused group query attention and (right) mediated attention. "*" refers to matrix multiplication, "$\sigma$" refers to non-linear transform, in this case softmax. In both cases, key and value are multiplied together before the query is applied.}
\label{fig:attention}
\label{fig:onecol}
\end{figure}

\textbf{Attention.} The self-attention blocks in DiT drive 40\% of compute requirements for S/2 (2.42 out of 6.07 GFLOPS). However, especially at late timesteps when the image is mostly noise, it is unclear if there is much for the attention blocks to "see", suggesting they might have excess capacity. In our base model, hidden layer $C = 384$ while token length $N = 256$, so we wanted to test modifications that did not just target the $\mathcal{O}(N^2)$ dependency.  We tested the following simplified attention blocks (see also Figure \ref{fig:attention}):
\begin{enumerate}
    \item \textbf{Shallow attention.} A naive simplification to halve the head dimension by setting the projection matrices $W_q$, $W_k$ and $W_v$ to shape $C \times C/2$. This halves the model's attention parameters and attention GFLOPS, though Big-O complexity is unchanged.
    \item \textbf{Mediated attention.} We introduced “mediator tokens” $T$ of dimension $h \times n \times d_h$, where $n << N$ (see Figure \ref{fig:attention}), as in \cite{pu2024efficientdiffusiontransformerstepwise}. $T$ was calculated by adaptive average pooling from $Q$. Attention was calculated in 2 steps, each with complexity $\mathcal{O} (NnC)$: 
        \begin{equation}
            V_\text{med} = \text{softmax}(TK^T / \sqrt{d})V
        \end{equation}
        \begin{equation}
            \text{Attn} = \text{softmax}(QT^T / \sqrt{d}) V_\text{med}
        \end{equation}
        
     We trialed $n=4$ and $n=16$. A depth-wise convolution of $V$ was added to the output as a simple proxy for local attention to features in $V$. However, we simplified the implementation by not adding bias terms to each mediator token multiplication.
    \item \textbf{Focused grouped-query / multi-query attention.} These methods are a novel approach combining the ideas of “focused attention” \cite{han2023flattentransformervisiontransformer} and “grouped query attention” \cite{ainslie2023gqatraininggeneralizedmultiquery}. “Focused attention” uses a polynomial ‘focusing’ function to project $Q$ and $K$ non-linearly, as a replacement for softmax non-linearity:
    \begin{equation}
        Q_p = \dfrac{||Q||}{||Q^p||}Q^p, \quad 
        K_p = \dfrac{||K||}{||K^p||}K^p
    \end{equation}
    We chose $p=3$ as in the reference. Because the softmax is eliminated and matrix multiplication is commutative, the first step in computation can be $K^TV$, as with mediated attention (see Figure \ref{fig:attention}). The resulting complexity is $\mathcal{O} (NC^2)$. In our case, because $C > N$, we combined focused attention with a grouped query approach (fewer $K$ and $V$ heads) to reduce the computations required. With grouping, the cost of the projection of $X$ to keys and values is reduced by a factor of $G$, and cost of $K^TV$ is reduced by $G^2$, for $G$ the grouping factor. We tested $G=3$ and $G=6$ (equivalent to multi-query attention).
\end{enumerate}

\textbf{Knowledge Distillation.} We used the DiT-S/2 model (32.7M parameters, trained for 200,000 steps) as the teacher and the significantly smaller DiT-XS/2 architecture (7.4M parameters, a 77\% reduction) as the student. Our distillation framework integrated two distinct loss components. First, we computed the standard diffusion loss ($L_{\text{diffusion}}$) using the denoising diffusion probabilistic model objective \cite{ho2020denoisingdiffusionprobabilisticmodels}: 
\begin{equation}
L_{\text{diffusion}} = \mathbb{E}_{x,\epsilon \sim \mathcal{N}(0,1),t} \left[ \| \epsilon - \epsilon_\theta(x_t, t, y) \|^2 \right]
\end{equation}

Second, we calculated a distillation loss ($L_{\text{KD}}$) measuring the mean squared error between the teacher and student model predictions \cite{hinton2015distillingknowledgeneuralnetwork}:
\begin{equation}
L_{\text{KD}} = \text{MSE}(\epsilon_{\text{student}}(x_t, t, y), \epsilon_{\text{teacher}}(x_t, t, y))
\end{equation}

We combined the two loss terms using a weighting parameter $\alpha \in [0, 1]$ to control the balance:
\begin{equation}
L_{\text{total}} = (1 - \alpha) \cdot L_{\text{diffusion}} + \alpha \cdot L_{\text{KD}}
\end{equation}

We set $\alpha = 0.3$ to moderately weight the teacher’s guidance. Lower values bias the model toward learning from raw data, while higher values shift emphasis toward reproducing the teacher's predictions.

During training, the teacher model was kept in evaluation mode with frozen weights, while only the student model was updated. For each batch, we computed both losses, combined them according to the weighting formula, and backpropagated through the student model. We trained using the AdamW optimizer with a learning rate of 1e-4 and weight decay of 0. After training, only the student model was retained for evaluation and inference; the teacher model was no longer required.

We loaded the pre-trained teacher model, disabled its gradients, and added the distillation loss calculation to the training loop. This approach allowed us to significantly reduce model size while attempting to preserve the generative capabilities of the larger model.

\textbf{Mixture of Experts.} Formally, a Mixture of Experts layer is defined as:
\begin{equation}
\mathrm{MoE}(x) \;=\; \sum_{i=1}^{E} g_{i}(x)\,e_{i}(x)
\end{equation}
where \(g_i(x)\) is a routing function over experts \(e_i(x)\). Each expert is implemented as a two-layer MLP with a GELU activation:
\begin{equation}
\mathrm{MLP}(x) \;=\; W_{2}\,\sigma_{\mathrm{gelu}}\bigl(W_{1}x\bigr)
\end{equation}
To further improve expert utilization, DiT-MoE introduces shared expert routing ensuring common knowledge is captured across experts, and an expert-level balance loss that regularizes routing load: 
\begin{equation}
\mathcal{L}_{\mathrm{balance}}
=\;\alpha \sum_{i=1}^{n}\frac{1}{K\,T}
\sum_{t=1}^{T} I(t,i)\;\frac{1}{\sum_{t'=1}^{T}P(t',i)}.
\end{equation}

In this set of experiments, we evaluate three Sparse Mixture‐of‐Experts variants on our DiT‐S/2 backbone (12 blocks, 384 hidden, patch=2, 6 heads) plus an extra‐small variant. DiT‐MoE‐S/2‐8E2A uses 8 experts with 2 active per token (MoE frequency=1), DiT‐MoE‐S/2‐4E1A uses 4 experts with 1 active (MoE frequency=2), and DiT‐MoE‐XS/2‐8E2A scales down to 6 blocks, 256 hidden, 4 heads with 8 experts and 2 active (MoE frequency=1).

We chose the 8E2A variant to match the original DiT-MoE\cite{fei2024scalingdiffusiontransformers16} small model, omitted the 16E2A variant to keep activated-parameter counts comparable to the base DiT-S/2, and added a half sized 4E1A variant to match the activated parameters with the base model’s 33M total parameters.

\begin{figure}
    \centering
    \includegraphics[width=\linewidth]{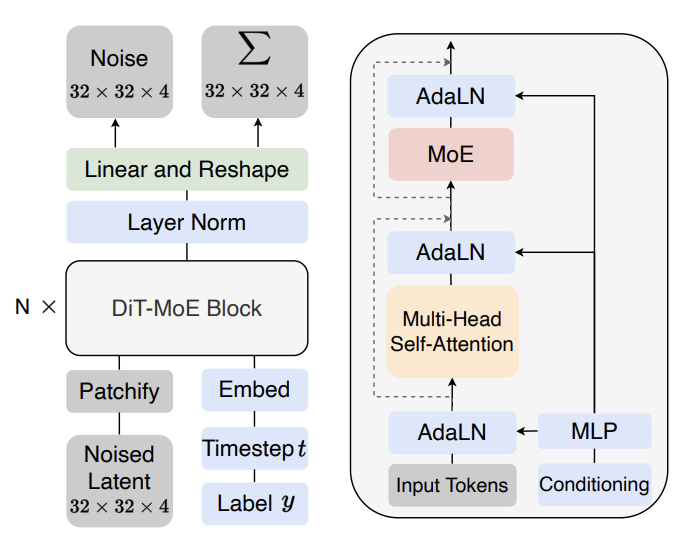}
    \caption{Overview of the DiT-MoE architecture. DiT-MoE extends the standard DiT by swapping each Transformer MLP for a sparse Mixture-of-Experts layer; the inset details how shared experts and token routing are integrated. \cite{fei2024scalingdiffusiontransformers16}}
    \label{fig:dit_moe_architecture}
\end{figure}

The resulting architecture (Figure~\ref{fig:dit_moe_architecture}) combines the scalability of sparse conditional computation with the performance of dense diffusion models.

\section{Experiments and Results}
\textbf{Evaluation methods.}
FID \cite{heusel2018ganstrainedtimescaleupdate} and spatial FID (sFID) \cite{nash2021generatingimagessparserepresentations} were computed using OpenAI's evaluator (\cite{dhariwal2021diffusionmodelsbeatgans}) applied to 10,000 images for the reference batch and sample batch. The reference batch was sampled from our reduced dataset of 200 classes with 50 images chosen at random from each class. Synthetic images were sampled with random class prompts using DiT's DDP image sampler \cite{peebles2023scalablediffusionmodelstransformers}. Both scores measure the distributional similarity between reference and sample image sets, with sFID more sensitive to spatial variability. Note that FID is known to change based on sample size \cite{jayasumana2024rethinkingfidbetterevaluation}, so our results are not comparable to the DiT results shown in \cite{peebles2023scalablediffusionmodelstransformers}. Other concerns with FID raised in \cite{jayasumana2024rethinkingfidbetterevaluation}, e.g., sometimes poor correlation with human-perceived quality, are reflected in our results, discussed in detail below.

\begin{figure}
    \setlength{\lineskip}{0pt}
    \centering
    \includegraphics[width=\linewidth]{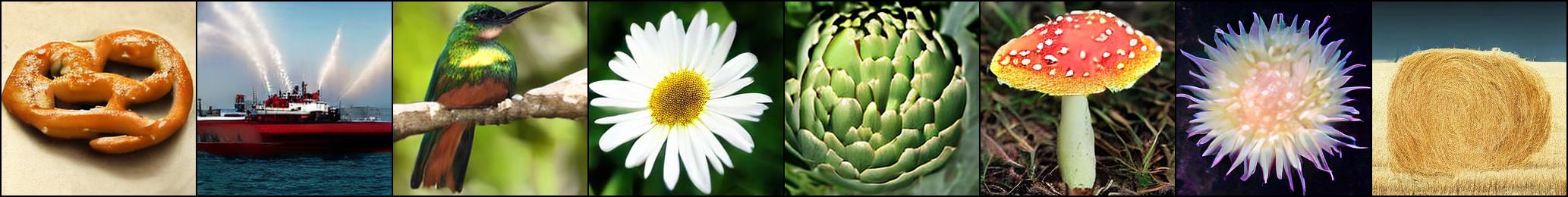}
    \includegraphics[width=\linewidth]{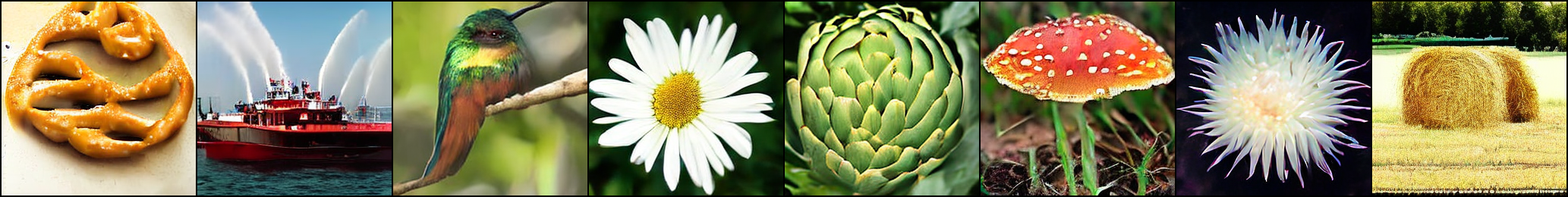}
    \includegraphics[width=\linewidth]{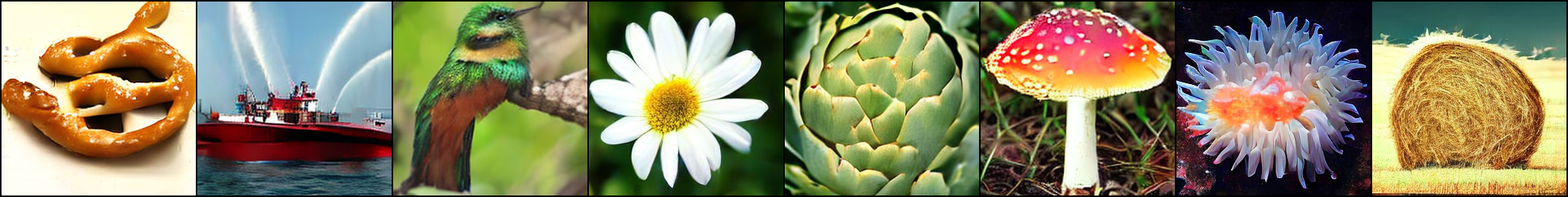}
    \includegraphics[width=\linewidth]{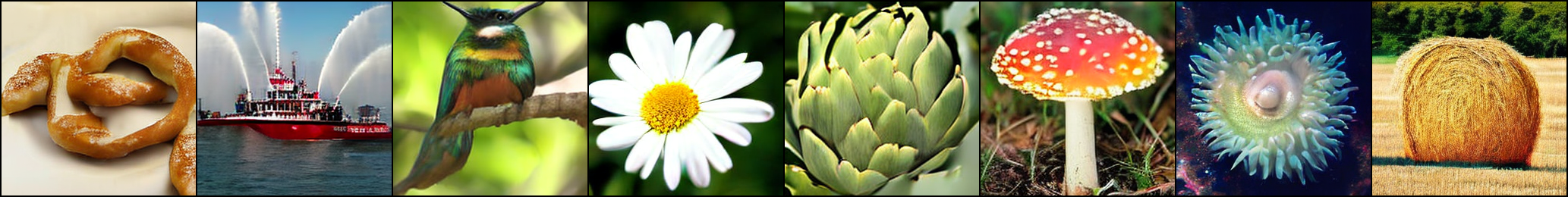}
    \includegraphics[width=\linewidth]{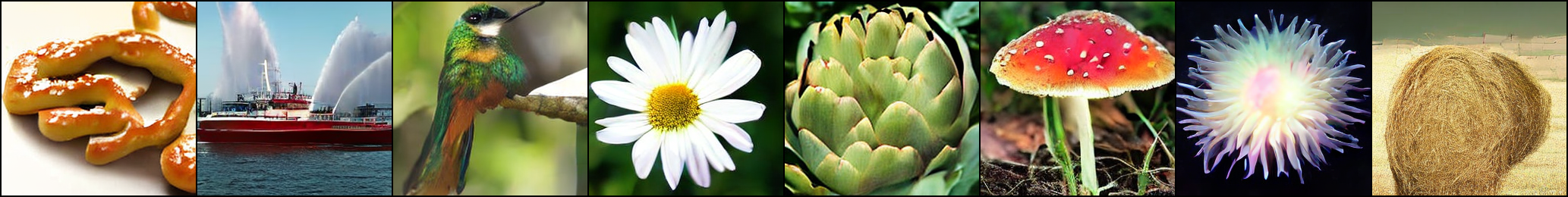}
    \includegraphics[width=\linewidth]{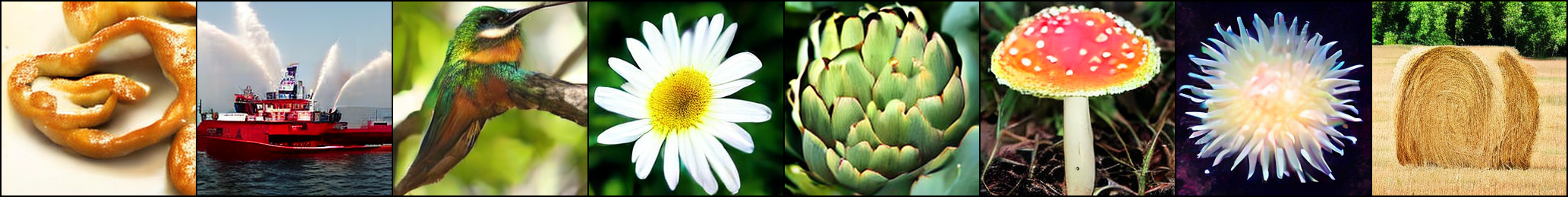}
    \includegraphics[width=\linewidth]{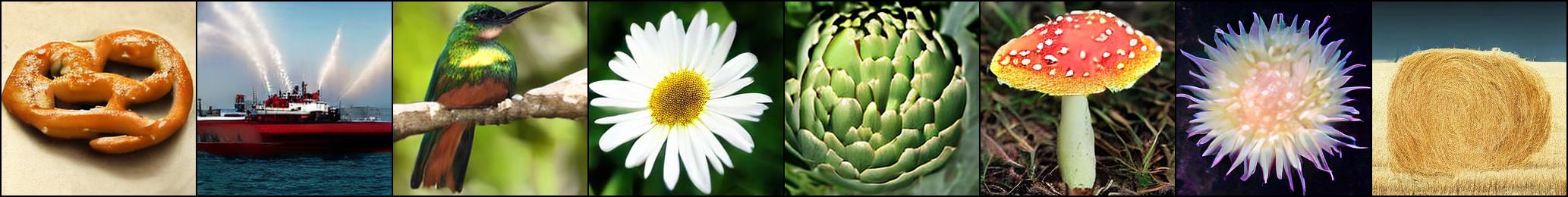}
    \includegraphics[width=\linewidth]{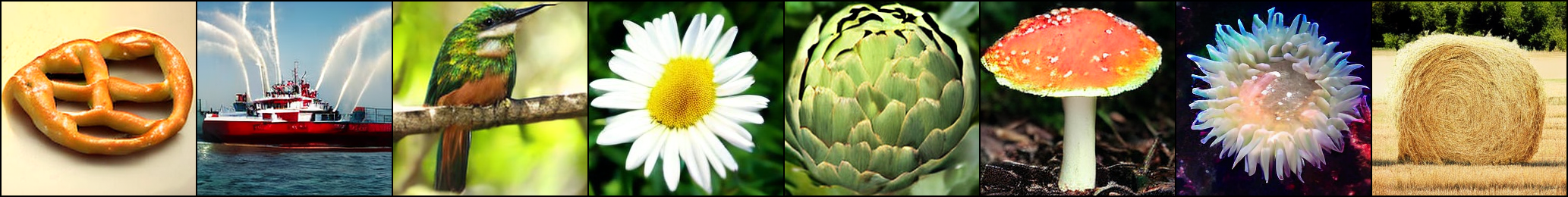}
    \includegraphics[width=\linewidth]{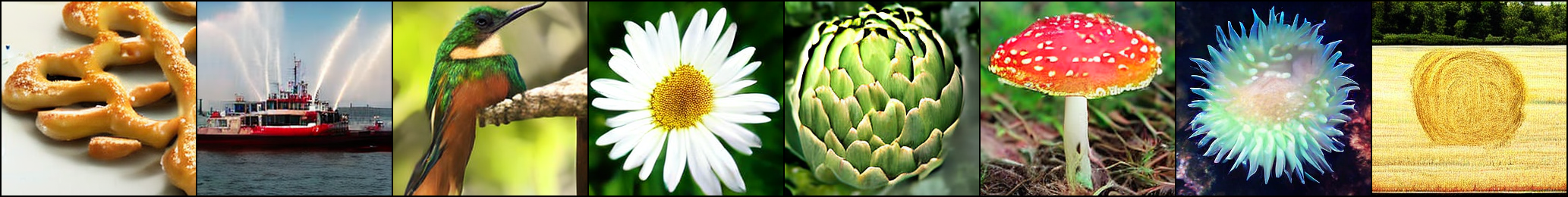}
    \includegraphics[width=\linewidth]{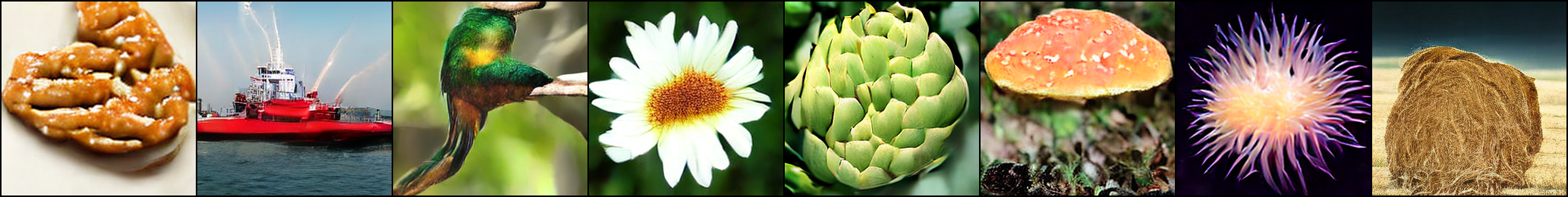}
    \includegraphics[width=\linewidth]{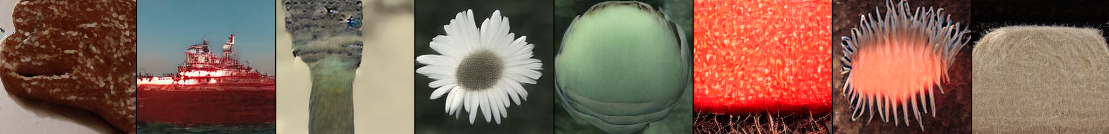}
    \caption{Selected samples (IDs) of pretzel (31), fireboat (164), jacamar (185), daisy (149), artichoke (1), mushroom (16), sea anemone (44), hay (179), generated with the same seed, CFG scale, and steps across all methods at 200K. Order: 1) S/2-base 2) S/2-shallow 3) S/2-med-4 4) S/2-med-16 5) S/2-fg-6 6) S/2-fg-3 7) Q-S/2 8) MoE-S/2-8E2A 9) MoE-S/2-4E1A 10) MoE-XS/2-8E2A 11) One pruned attention head.}
    \label{fig:qual_samples}
    
\end{figure}

Profiling used a standardized script that compiled each model for max-autotune optimization, included warm-up, and measured throughput over 50 inference iterations. Peak memory was tracked via \texttt{torch.cuda.max\_memory\_allocated}, and FLOPs were counted using fvcore. Reproducing FLOP counts from prior work required removing transformer block wrapper code~\cite{peebles2023scalablediffusionmodelstransformers}. For accurate FLOP counting, attention blocks were modified during profiling to bypass PyTorch's CUDA-optimized dot product attention function. Image quality was evaluated qualitatively using 8 pre-defined class prompts. Results are shown in Table \ref{tab:performance} and Figure \ref{fig:qual_samples}. 

\begin{table*}
\begin{center}
\begin{tabular}{|l|c|c|c|c|c|c|}
\hline
Model & Params (M) & Peak memory (GB) & FID-10K & sFID-10K & GFLOPS & Throughput (it / s) \\
\hline\hline
S/2-base  & 32.7 & 0.267 & 22.9 & 38.7 & 6.07 &  9.07 \\
S/4-base  & 32.7 & 0.447 & - & - & 1.42 &  7.4 \\
XS/2-base  & 7.5 & 0.389 & 110.5 & 44.6 & 1.42 &  8.2 \\
XS/4-base  & 7.5 & 0.431 & - & - & 0.32 &  7.69 \\

\textbf{Quantization} & & & & & & \\
Q-S/2 & 32.7 & 0.24 {\scriptsize (–10.1\%)} & 55.7 & 36.3 & 6.07 & 6.91 {\scriptsize (-23.8\%)} \\
Q-S/4 & 32.7 & 0.277 {\scriptsize (-38\%)} & - & - & 1.42 & 5.66 {\scriptsize (-23.5\%)} \\
Q-XS/2 & 7.5 & 0.193 {\scriptsize (-50.4\%)} & - & - & 1.42 & 6.81 {\scriptsize (-17\%)} \\
Q-XS/4 & 7.5 & 0.14 {\scriptsize (-67.2\%)} & - & - & 0.32 & 6.08 {\scriptsize (-20.9\%)} \\

\textbf{Attention} & & & & & & \\
S/2-shallow  & 29.1 & 0.238 & 24.9 & 39.2 & 4.86 & 8.88  \\
S/2-med-4 & 32.7 & 0.253 & 21.5 & 39.0 & 5.49 &  7.70 \\
S/2-med-16 & 32.7 & 0.255 & 21.4 & 39.7 & 5.55 &  7.97 \\
S/2-fg-6 & 29.7 & 0.233 & 23.6 & 39.6 & 4.80 &  8.53 \\
S/2-fg-3 & 30.3 & 0.239 & 22.3 & 39.2 & 4.96 &  7.87 \\

\textbf{Knowledge Distillation} & & & & & & \\
KD-XS/2 & 7.5 & 0.389 & 110.8 & 45.1 & 1.42 & 8.2 \\

\textbf{Mixture of Experts} & & & & & & \\
MoE-S/2-8E2A & 131.9 [47] & 0.918 & 42.9 & 35.3 & 9.70 & 4.11 \\
MoE-S/2-4E1A & 53.9 [32] & 0.469 & 56.5 & 36.7 & 6.07 & 5.98 \\
MoE-XS/2-8E2A & 29.6 [10] & 0.280 & 98.9 & 42.9 & 2.23 & 11.02 \\

\hline
\end{tabular}
\end{center}
\caption{Quantitative performance comparison of model memory (params, peak), image quality (FID, sFID), and compute (GFLOPS, throughput). Brackets indicate the number of activated parameters activated for MoE models.}
\label{tab:performance}
\end{table*}

\begin{table*}
\begin{center}
\begin{tabular}{|c|c|c|c|c|c|c|c|c|c|c}
\hline
S/2-shallow & S/2-med-4 & S/2-med-16 & S/2-fg-6 & S/2-fg-3 & MoE-S/2-8E2A & MoE-S/2-4E1A & MoE-XS/2-8E2A \\
\hline\hline
0.1584 & 0.1556 & 0.1552 & 0.1548 & 0.1556 & 0.1444 & 0.1524 & 0.1564 \\

\hline
\end{tabular}
\end{center}
\caption{Final training losses (200k steps) for each optimization strategy.}
\label{tab:trainingloss}
\end{table*}

\begin{figure}
    \centering
    \includegraphics[width=\linewidth]{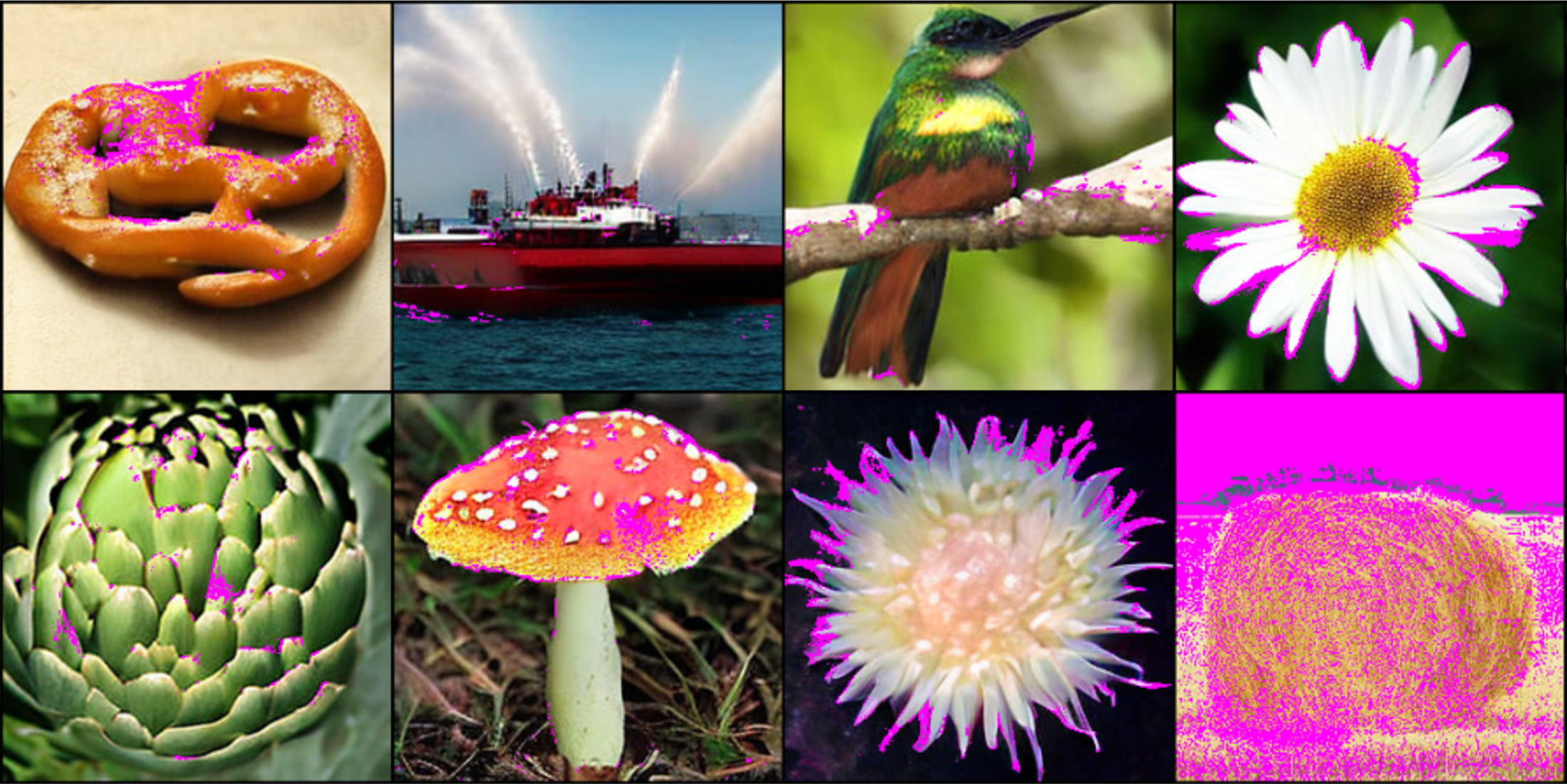}
    \caption{Pixel-wise differences between DiT-S/2 baseline and quantized outputs; magenta regions shows where the quantized samples deviate from the baseline.}
    \label{fig:quant_samples}
\end{figure}

\textbf{Pruning.} Even a single pruned attention head, as seen in the bottom row of Figure \ref{fig:qual_samples}, introduces significant artifacts and deterioration in the model's ability to represent fine-grained details relative to the baseline. This suggests the criticality of the attention heads in the DiT model's representational power, particularly with a relatively small model as S/2, having only six attention heads. While pruning successfully reduces memory footprint and computational requirements, this degradation in visual fidelity reveals a clear quality-efficiency tradeoff that needs to be considered when deploying pruned models.

\textbf{Quantization}. We evaluated four DiT model variants (S/2, S/4, XS/2, XS/4) without any fine-tuning or retraining after quantization. Quantization reduced peak memory usage by 10\%-68\% and compressed model size by approximately 73\% across all variants, shrinking S/2 and S/4 models from 131MB to 34MB and XS/2 and XS/4 models from 30MB to 8MB. This level of compression aligns with expectations for float32-to-int8 quantization, which yields a $4\times$ reduction in per-weight storage.

However, inference throughput decreased by 17\%-24\%, likely due to the experimental nature of TorchAO’s int8 CUDA kernels, which are not fully optimized. Despite the unexpected slowdown in throughput, quantization achieved its primary objective of substantial memory and model compression with minimal implementation overhead.

The FID and sFID of the quantized DiT-S/2 model were 55.70 and 36.28 respectively, higher than the baseline S/2 (Table \ref{tab:performance}). Figure \ref{fig:quant_samples} shows minimal visual differences, suggesting quantization achieved effective compression without noticeable quality loss. Notably, the majority of pixel-level deviations appear in regions with high-frequency textures or noise-like details, whereas smooth or low-frequency areas remain nearly identical between the baseline and quantized outputs. This suggests that quantization noise has a greater impact on fine-grained visual detail but is largely imperceptible in natural, low-complexity regions.

\textbf{Attention models}. Five attention models were trained from scratch for 200,000 steps. In each model all attention blocks were replaced with shallow attention, mediated attention with \textit{n}=4 or 16, or focused group/multi-query attention with \textit{G}=3 or 6 respectively, enabling us to observe performance trends if present. Although uptraining is commonly used (e.g., \cite{ainslie2023gqatraininggeneralizedmultiquery}), we trained from scratch to better understand the tradeoffs between model compression, compute reduction and image quality. 

Table \ref{tab:performance} shows that each attention variant has similar FID and sFID to the base S/2. Shallow attention reduces compute demand (GFLOPS) by 20\% but with somewhat worse FID. Using mediated attention and focused group query attention blocks maintains inference quality while reducing model memory requirements and compute.  Mediated attention models see both reduced memory and compute (by a modest 8.5-9.5\%) and improved FID over base S/2. Focused multi-query attention (fg-6) aggressively trims the model with somewhat worse FID, but the focused group query attention with 2 key-value heads (fg-3) achieves similar FID to base S/2 while reducing GFLOPS 18\% and with 2 million fewer parameters. Qualitatively, sampled generated images (see Figure \ref{fig:qual_samples}) look somewhat different from the base. FID might not be an accurate reflection of human-perceived quality.

Contrasting with the results from attention head pruning, these results suggest that there maybe 'excess capacity' in the attention blocks of S/2, with even a naive halving of attention parameters (i.e. shallow attention) producing a reasonable output, but that having multiple layers of attention matters. Note that none of the attention models achieved faster inference throughput than the base S/2, likely because the base model uses a CUDA-optimized attention function which was not used for the modified attention calculations. It appears that better hardware performance dominates the number of calculations required.

\textbf{Knowledge Distillation.} We trained two DiT-XS/2 models to assess distillation: one with a DiT-S/2 teacher (200K steps) and one without. The distilled model slightly underperformed the non-distilled baseline (Table \ref{tab:performance}), suggesting ineffective knowledge transfer for this student-teacher pair. Both DiT-XS/2 variants showed substantially higher FID and sFID scores than the DiT-S/2 base model, indicating a degradation in generative quality due to reduced model capacity. While they achieved the expected throughput gains, this came at the cost of generative performance.

The 77\% reduction in parameters likely introduced a capacity gap too large for effective knowledge transfer from the DiT-S/2 teacher, despite a distillation loss weight of $\alpha = 0.3$. Future work could explore hyperparameter sweeps across different $\alpha$ values, or apply distillation to student models like DiT-S/4 or S/8, which reduce compute via larger patch sizes while retaining the same parameter count. Alternatively, using wider or deeper student models could narrow the capacity gap and enable better alignment with the teacher.

\textbf{Mixture of Experts.} All MoE variants (Table~\ref{tab:performance}) show significantly worse FID scores than the DiT-S/2 baseline, with our smallest variant (MoE-XS/2-8E2A) quadrupling the S/2-base's 22.9 FID-10K to 98.9. This degradation likely reflects imbalanced expert utilization despite applying a balance loss. Interestingly, spatial FID improves slightly for MoE-S/2-8E2A (35.3 vs. 38.7), suggesting better local structure capture despite poor global statistics (FID).

MoE layers increase parameter count (32.7M to 131.9M for S/2-8E2A), peak memory, and GFLOPS (6.07 to 9.70), reducing throughput from 9.07 to 4.11 it/s on S/2-8E2A. The XS variant, with only 29.6M total (10M "activated") parameters, runs faster (11.02 it/s) at very low GFLOPS (2.23). Our MoE modifications successfully increase capacity and arithmetic intensity with sensible runtime scaling, but require further tuning (e.g. routing balance or expert capacity) to recover or surpass baseline generative performance.

In Figure \ref{fig:qual_samples}, larger‐capacity MoE variants render richer backgrounds and finer details, with more coherent textures compared to the S/2 baseline, reflecting their higher parameter counts. Conversely, the XS/2-8E2A model, despite its sparse “activated” footprint, appears under-provisioned. In this case, the bird is distorted and the mushroom stem is missing, indicating that its smaller expert capacity struggles to capture global structure.



\clearpage
\twocolumn
{\small
\bibliographystyle{ieee_fullname}
\bibliography{egbib}
}

\end{document}